\documentclass{article}

\usepackage{PRIMEarxiv}

\usepackage[utf8]{inputenc}
\usepackage[T1]{fontenc}
\usepackage{microtype}
\usepackage{inconsolata}
\usepackage{graphicx}
\usepackage{booktabs}
\usepackage{multirow}
\usepackage{amsmath}
\usepackage{xcolor}
\usepackage{listings}
\usepackage{enumitem}
\usepackage{caption}
\usepackage{tikz}
\usepackage{url}
\usetikzlibrary{arrows.meta, positioning}

\usepackage[numbers]{natbib}

\definecolor{hallbg}{HTML}{dce8f5}
\definecolor{hallbd}{HTML}{3b6ea8}
\definecolor{rwdbg}{HTML}{fde8d8}
\definecolor{rwdbd}{HTML}{c4793b}
\definecolor{accbg}{HTML}{ddf0e1}
\definecolor{accbd}{HTML}{3a7d44}
\definecolor{iobg}{HTML}{f2f2f2}
\definecolor{iobd}{HTML}{888888}
\definecolor{passclr}{HTML}{3a7d44}
\definecolor{rejclr}{HTML}{b03030}

\graphicspath{{figures/}}

\title{\includegraphics[width=1\linewidth]{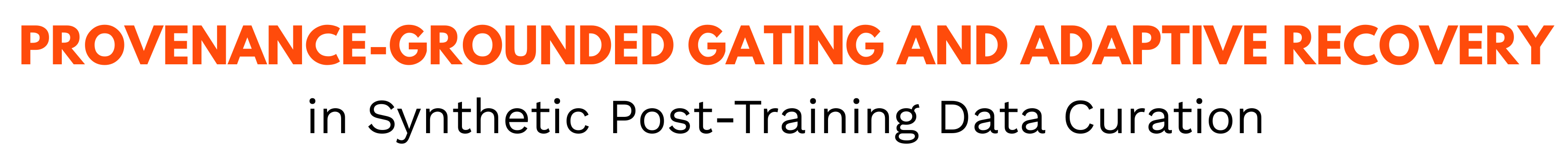}}

\author{
  Soham Bhattacharjee, Karun Sharma, \\
  Vinay Kumar Sankarapu, Pratinav Seth\\ 
  \affiliation{Lexsi Labs} \\
  \texttt{soham.bhattacharjee@lexsi.ai}
}

\setabstract{%
Synthetic post-training pipelines commonly filter generated samples with reward models or holistic LLM judges, yet two practices remain rarely examined together: whether the filtering signal is grounded in the source evidence that induced each generation, and whether rejected samples can be systematically recovered rather than permanently discarded. We present a controlled study of both questions across gate configurations, recovery strategies, and generator scales, using adversarially injected corpora to provide ground-truth failure labels. We find that exact source provenance improves faithfulness gating for stronger judges, that hallucination and reward gates reject largely disjoint sample populations making both necessary, and that an adaptive recovery pipeline combining failure diagnosis with targeted regeneration achieves higher yield, recovery rate, and injection recall than naive resampling. Downstream fine-tuning quality is driven primarily by generator scale, with filtration and recovery conditions contributing meaningfully but secondarily. 
}

\setkeywords{LLM-as-judge, synthetic data curation, hallucination detection, data quality, provenance}

\runningtitle{Provenance-Grounded Gating and Adaptive Recovery}

\begin{document}
\maketitle

\section{Introduction}
\label{sec:intro}

Synthetic post-training pipelines filter generated samples with
LLM-as-judge quality gates before fine-tuning. Judges in this setting
typically follow preference rubrics covering helpfulness, instruction
following, and truthfulness \citep{cui2023ultrafeedback}, but these
rubrics do not verify whether a generated response is supported by
the source passage that induced it. Source-grounded evaluation
frameworks such as G-Eval \citep{liu2023geval} and CheckEval
\citep{lee-etal-2025-checkeval} derive their criteria directly from a source
chunk and offer a more targeted faithfulness signal, yet none of the
major synthetic data curation libraries, including distilabel
\citep{distilabel} and AgentInstruct \citep{agentinstruct}, preserve
an explicit provenance record linking each generated sample back to
its generation source. Without this record, faithfulness gating must
operate on post-hoc retrieved evidence, which approximates rather
than observes the original evidence relation.

A second gap concerns rejected samples. Existing pipelines treat
rejection as terminal, discarding failures and spending generation
budget on fresh samples. The rejection sampling literature, including
RAFT \citep{dong2023raft} and ReST\textsuperscript{EM}
\citep{singh2024beyond}, shows that regenerating from the same prompt
under a quality filter substantially improves yield. These methods
target iterative model self-improvement under a single global
acceptance criterion, however, and whether structured failure
diagnosis and targeted repair offers meaningful gains over naive
regeneration in a source-grounded curation setting is an open
question.

We study both problems in a controlled ablation, varying gate
configuration, recovery strategy, and generator scale while holding
corpus and judge family fixed. We find that exact-provenance gating
outperforms both reward-only filtering and post-hoc retrieved evidence
on faithfulness detection, that hallucination and reward gates reject
largely disjoint failure populations and are thus both necessary, and
that adaptive diagnose-and-repair outperforms naive regeneration on
yield, recovery rate, and injection recall. Downstream fine-tuning
quality is driven primarily by generator scale, with filtration and
recovery conditions contributing meaningfully but secondarily.
\newpage
\textbf{Contributions:}
\begin{enumerate}
    \item A provenance-grounded gating study showing that exact source evidence improves faithfulness detection for stronger judges, with reward-only filtering and post-hoc retrieval degrading gate quality in complementary ways.
    \item A controlled recovery ablation showing that adaptive diagnose-and-repair consistently outperforms naive regeneration in yield, recovery rate, and injection recall.
    \item A clean held-out evaluation corpus for source-grounded QA, released to support reproducible downstream evaluation of synthetic data curation pipelines.\footnote{\url{https://huggingface.co/datasets/Lexsi/provenance-grounded-synthetic-qa}}
\end{enumerate}

\section{Methodology}
\label{sec:method}

\textbf{Provenance-preserving generation.}
Each sample is generated from a source chunk $c$ drawn from the
corpus, and a provenance record linking the sample to $c$ is attached
at generation time. This record is append-only and persists through
all downstream pipeline stages, making the exact generation evidence
available to the hallucination gate without retrieval.

\textbf{Gates.}
The \emph{HallucinationGate} performs structured claim verification
against the preserved source chunk ($\tau_{\text{hall}} = 0.8$),
directly testing whether the generated response is entailed by the
evidence that induced it. The \emph{RewardGate} scores
instruction-output quality following a preference rubric without
source access ($\tau_{\text{reward}} = 0.7$). A sample must clear
both gates to be accepted.

\textbf{Adaptive recovery (primary).}
On hallucination failure, a \emph{DiagnosticProbe} diagnoses the
failure mode and applies a targeted configuration patch, such as
lowering generation temperature or expanding context, before
regenerating the sample. Probe-recovered samples re-enter the reward
gate. On reward failure, a \emph{RewardRefiner} rewrites the output
for quality without altering the underlying claims. Both recovery
steps operate against the same preserved provenance record.

\textbf{Naive retry (ablation).}
On any gate failure, the generator produces a fresh response from the
original prompt without diagnosis or repair. The new response must clear both gates with up to five retries before the sample is permanently rejected.

\begin{figure}[pt]
  \centering
  \includegraphics[width=0.5\textwidth]{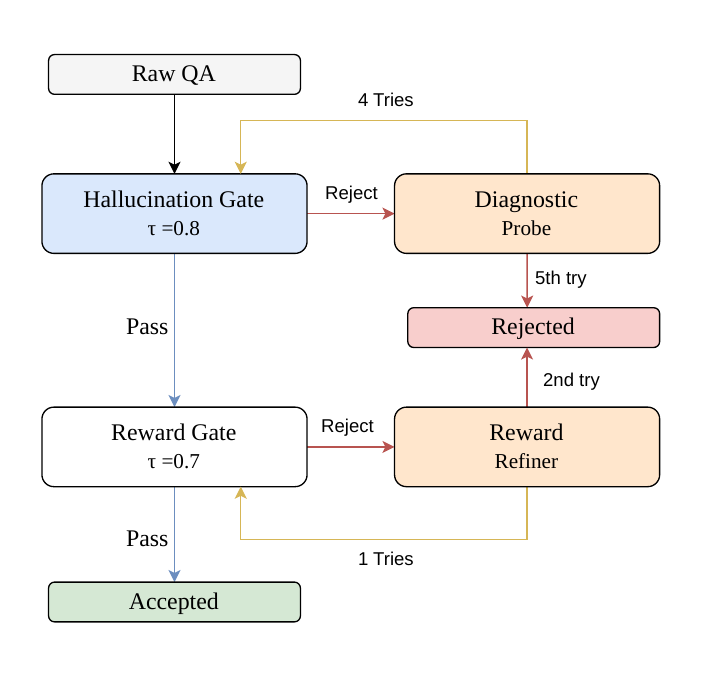}
  \caption{Adaptive recovery pipeline. Hall-rejected samples enter the
    DiagnosticProbe for up to 4 targeted repair attempts (blue path);
    reward-rejected samples enter the RewardRefiner for one rewrite
    (orange path). Samples clearing both gates are accepted.}
  \label{fig:pipelines}
\end{figure}

\section{Experimental Setup}
\label{sec:setup}

\textbf{Source corpus.}
The corpus spans three domains: $1{,}500$ source chunks ($500$ per
domain) from CUAD~\citep{cuad} commercial contracts,
PubMedQA~\citep{jin2019pubmedqa} biomedical abstracts, and English
Wikipedia 2023.\footnote{\url{https://huggingface.co/datasets/wikimedia/wikipedia}}
Each chunk seeds three base QA pairs per training variant; across two training variants and post-generation filtering of malformed outputs, this yields approximately 8,000–8,500 raw candidates per generator.

\textbf{Adversarial injections.}
Four failure types are injected into approximately $20\%$ of samples:
\emph{contradicts\_source}, \emph{parametric\_drift},
\emph{domain\_mismatch}, and \emph{instruction\_quality},
providing ground-truth labels for gate recall value of rejecting
these samples and recovering them.

\textbf{Generators and judges.}
Generators: Qwen3-1.7B, 4B, 8B~\citep{qwen3_techreport}. Three
judge sizes are used across experiments: Qwen3-14B, Qwen3.6-27B-FP8,
and Qwen3.6-35B-A3B. Provenance gating experiments are evaluated
across all three judges; adaptive recovery experiments use the 14B
and 35B judges; the recovery ablation uses a single judge. All
models run in non-thinking mode via vLLM.

\textbf{Benchmarks and metrics.}
Provenance gating quality is evaluated on
FaithDial~\citep{faithdial} ($n{=}3{,}507$) against
ground-truth faithfulness labels. Downstream evaluation uses a
held-out test set of $1{,}400$ instructions, generated independently
of all training corpora and fixed before any pipeline experiments
began. Metrics: ROUGE-L, BERTScore~$F_1$, and Faithfulness
(BERTScore~$F_1$ between model output and source chunk).

\textbf{Fine-tuning.}
Qwen3-4B base, LoRA $(r{=}16,\ \alpha{=}32)$ via
Unsloth~\citep{unsloth,hu2021lora}, early stopping (patience~$=2$,
up to 4 epochs). Full hyperparameters in
Table~\ref{tab:hparams} (Appendix~\ref{app:hyperparams}).

\section{Results and Analysis}
\label{sec:results}
\begin{figure}[pt]
  \centering
  \includegraphics[width=0.65\textwidth]{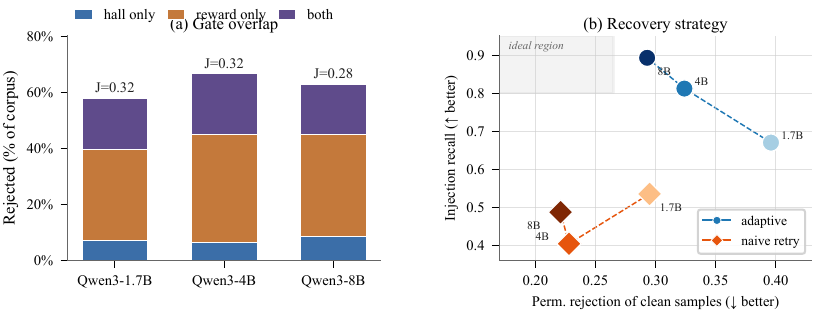}
  \caption{\textbf{(a)}~Rejection-set overlap per generator as \% of
    corpus (35B judge; Appendix~\ref{app:gate_overlap_14b} for 14B
    numbers). Both hall-only and reward-only segments are substantial;
    Jaccard $\in[0.23, 0.32]$, confirming the two gates target
    structurally different failure modes.
    \textbf{(b)}~Recall-rejection tradeoff (shaded = better; ideal:
    lower-right). Adaptive improves with generator size (light to dark
    blue); by 8B ($29.3\%$ reject, $89.3\%$ recall) it achieves lower
    rejection \emph{and} higher recall than naive retry on 1.7B
    ($29.5\%$, $53.5\%$; diamond), fully dominating on both axes.}
  \label{fig:gates}
\end{figure}

\begin{table}[pt]
  \centering
  \footnotesize
  \caption{Gate method comparison on FaithDial ($n{=}3{,}507$).
           Exact-provenance gating achieves the highest $F_1$ at 27B
           and 35B; retrieved provenance leads marginally at 14B.
           Reward-only scoring catastrophically over-rejects faithful
           samples at every judge size. Full P/R results in
           Appendix~\ref{app:exp1a}.}
  \label{tab:exp1a}
  \vspace{4pt}
  \setlength{\tabcolsep}{5pt}
  \begin{tabular}{lccc}
    \toprule
    \textbf{Gate configuration} & \textbf{14B} $F_1$ & \textbf{27B} $F_1$ & \textbf{35B} $F_1$ \\
    \midrule
    Exact provenance (ours) & 0.494          & \textbf{0.594}          & \textbf{0.614} \\
    Retrieved provenance    & \textbf{0.514} & 0.573          & 0.590          \\
    Oracle holistic         & 0.489          & 0.569          & 0.599          \\
    Reward only             & 0.291          & 0.185          & 0.184          \\
    \bottomrule
  \end{tabular}
\end{table}

We evaluate the pipeline along three axes: whether exact-provenance
gating detects faithfulness failures missed by reward-only filtering
and whether the two gates are complementary, whether adaptive
diagnosis-and-repair outperforms naive regeneration as a recovery
strategy, and whether curation choices affect downstream fine-tuning
quality independently of generator scale (Section~\ref{sec:downstream}).

\subsection{Gate Characterization}
\label{sec:gates}

We evaluate four gate configurations on FaithDial ($n{=}3{,}507$);
results are in Table~\ref{tab:exp1a}. Exact-provenance gating achieves the highest $F_1$ at 27B and 35B,
reaching $0.614$ at 35B, while retrieved provenance leads marginally
at 14B ($0.514$ vs. $0.494$). Overall, post-hoc retrieval trails exact
provenance for stronger judges, and oracle holistic scoring reaches
$0.599$ at 35B. Reward-only scoring is unusable as
a faithfulness filter: only $14$-$45\%$ of genuinely faithful samples
pass depending on judge strength, making it a source of systematic
over-rejection rather than quality control. The provenance signal
transfers across judge families under both G-Eval and CheckEval
scoring (Appendix~\ref{app:exp1d}), confirming the finding is not an artefact of a particular judge model.

Figure~\ref{fig:gates}(a) breaks down rejection sets when both gates run on the unfiltered generated corpus. Jaccard overlap sits in
$[0.23, 0.32]$ across all generators and both judges
(Appendix~\ref{app:gate_overlap_14b}), confirming the two gates target structurally different failure modes. The split is not random: \emph{contradicts\_source} and \emph{domain\_mismatch} are caught reliably ($82$-$90\%$ recall for the 4B and 8B generators), while \emph{instruction\_quality} resists both gates
($24$-$42\%$ recall; Table~\ref{tab:gate_recall},
Appendix~\ref{app:injection_recall}). Neither gate catches what the other misses; both are necessary.

\begin{table}[t]
  \centering
  \footnotesize
  \caption{Gate rejection-set overlap on the generated corpus (35B
           judge). Hall-only and reward-only segments are both
           substantial; Jaccard $\in[0.28, 0.32]$ confirms the two
           gates target structurally different failure modes.
           14B numbers in Appendix~\ref{app:gate_overlap_14b}.}
  \label{tab:exp1b}
  \vspace{4pt}
  \setlength{\tabcolsep}{4pt}
  \begin{tabular}{lcccc}
    \toprule
    & & \multicolumn{3}{c}{\textit{Rejected (\% of corpus)}} \\
    \cmidrule(lr){3-5}
    \textbf{Gen.} & \textbf{Jaccard} & \textbf{Hall} & \textbf{Reward} & \textbf{Both} \\
    \midrule
    1.7B & 0.317 & 7.3 & 32.3 & 18.4 \\
    4B   & 0.324 & 6.5 & 38.6 & 21.6 \\
    8B   & 0.284 & 8.5 & 36.6 & 17.8 \\
    \bottomrule
  \end{tabular}
\end{table}

\subsection{Recovery: Structured Repair Outperforms Naive Retry}
\label{sec:recovery}

Structured adaptive repair outperforms naive retry on total yield,
pass rate, and injection recall across all three generator sizes.
The full stage-by-stage breakdown including the $+42\%$ yield gain
over hard filtering is in Appendix~\ref{app:adaptive}.

\begin{table}[t]
  \centering
  \footnotesize
  \caption{Adaptive recovery vs.\ naive retry (14B judge). Adaptive
         leads on pass rate, recovery rate, and injection recall at
         every generator size, with the injection recall gap widening
         substantially with scale. Naive retry permanently discards
         fewer clean samples (lower uninjected rejection), but
         adaptive recovers enough of its initial rejects to end with
         higher total yield despite this disadvantage.}
  \label{tab:exp2b}
  \vspace{4pt}
  \setlength{\tabcolsep}{3.5pt}
  \begin{tabular}{lrrrrrr}
    \toprule
    & \multicolumn{3}{c}{\textbf{Adaptive}} & \multicolumn{3}{c}{\textbf{Naive retry}} \\
    \cmidrule(lr){2-4}\cmidrule(lr){5-7}
    \textbf{Metric} & \textbf{1.7B} & \textbf{4B} & \textbf{8B}
                    & \textbf{1.7B} & \textbf{4B} & \textbf{8B} \\
    \midrule
    Pass rate (\%)     & 71.6 & 85.8 & \textbf{88.5} & 68.4 & 73.6 & 72.2 \\
    Recovery rate (\%) & 32.5 & 66.3 & \textbf{72.7} & 24.9 & 36.4 & 34.6 \\
    Inj.\ recall (\%)  & 67.0 & 81.2 & \textbf{89.3} & 53.5 & 40.4 & 48.7 \\
    Uninj.\ reject (\%) & 39.6 & 32.4 & 29.3 & \textbf{29.5} & \textbf{22.8} & \textbf{22.1} \\
    \bottomrule
  \end{tabular}
\end{table}

Adaptive accepts more samples and recovers a larger share of total
rejects at every generator size, with the recovery rate advantage
growing from $+7.6$ points at 1.7B to $+38.1$ points at 8B. The
injection recall gap is the starkest signal: $89.3\%$ vs $48.7\%$
at 8B, nearly double, with the widest within-type gap on
\emph{instruction\_quality} ($74.4\%$ vs $51.3\%$) and
\emph{parametric\_drift} ($71.7\%$ vs $58.1\%$) at 1.7B
(Table~\ref{tab:gate_recall}, Appendix~\ref{app:injection_recall}).
Naive retry permanently discards fewer samples from the non-injected
pool, which we term the natural rejection rate. We note that
non-injected samples are not guaranteed to be high quality; they
simply did not receive a controlled adversarial perturbation, and
natural gate failures among them may reflect genuine faithfulness
or quality deficiencies rather than over-rejection. Adaptive's
higher natural rejection rate therefore does not straightforwardly
imply worse precision (see Section~\ref{app:discussion}).
Figure~\ref{fig:gates}(b) shows the recall-rejection tradeoff:
adaptive improves on both axes as generator size increases, and by
8B fully dominates naive retry on both dimensions simultaneously.

\subsection{Downstream: Generator Size Dominates}
\label{sec:downstream}

\begin{figure}[pt]
  \centering
  \includegraphics[width=0.55\textwidth,height=0.27\textwidth]{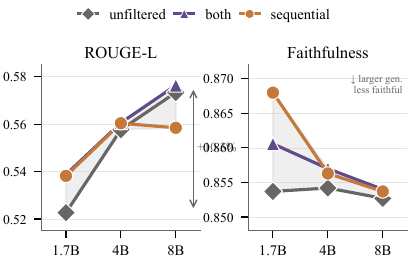}
  \caption{Downstream metrics by generator and condition. Each line
    shows one filtration condition across generator sizes; the shaded
    band spans the within-generator spread. The steep cross-generator
    slope ($+6.6\%$ ROUGE-L, $1.7\text{B}{\to}8\text{B}$) dwarfs the
    within-generator spread, confirming generator scale dominates
    downstream quality.}
  \label{fig:downstream}
\end{figure}

\begin{table}[t]
  \centering
  \footnotesize
  \caption{Downstream fine-tuning (Qwen3-4B base, LoRA, $n{=}1{,}400$).
    Three representative conditions are shown; full results across all
    five conditions in Table~\ref{tab:exp2c_full}
    (Appendix~\ref{app:adaptive}).
    Generator scale drives ROUGE-L ($+6.6\%$,
    $1.7\text{B}{\to}8\text{B}$); within-generator spread across all
    conditions is at most $0.016$ (1.7B), $0.007$ (4B), $0.018$ (8B).
    Bold = best per generator per metric.}
  \label{tab:exp2c}
  \vspace{4pt}
  \setlength{\tabcolsep}{3.5pt}
  \begin{tabular}{llccc}
    \toprule
    \textbf{Gen.} & \textbf{Condition} & \textbf{R-L} & \textbf{BS-F1} & \textbf{Faith.} \\
    \midrule
    1.7B & unfiltered  & 0.5227 & 0.9059 & 0.8537 \\
    1.7B & both        & \textbf{0.5388} & 0.9112 & 0.8606 \\
    1.7B & sequential  & 0.5382 & \textbf{0.9113} & \textbf{0.8680} \\
    \midrule
    4B   & unfiltered  & 0.5577 & 0.9185 & 0.8542 \\
    4B   & both        & \textbf{0.5606} & 0.9197 & \textbf{0.8570} \\
    4B   & sequential  & 0.5604 & \textbf{0.9223} & 0.8563 \\
    \midrule
    8B   & unfiltered  & 0.5734 & 0.9187 & 0.8527 \\
    8B   & both        & \textbf{0.5764} & \textbf{0.9191} & \textbf{0.8540} \\
    8B   & sequential  & 0.5585 & 0.9189 & 0.8537 \\
    \bottomrule
  \end{tabular}
\end{table}

Generator scale is the dominant signal: ROUGE-L rises from $0.523$
(1.7B, unfiltered) to $0.576$ (8B, both-filtered), a cross-generator
gain of $+6.6\%$ that dwarfs any within-generator spread. The
practical contribution of filtration shrinks with generator capacity:
the gain from unfiltered to best-filtered is $+0.016$ ROUGE-L for
1.7B but only $+0.003$ for 4B and 8B, a fivefold reduction. For the
two larger generators, all conditions including unfiltered cluster
within $0.007$ of each other, indicating that stronger generators
produce data of sufficient baseline quality that curation adds
marginal downstream value.

Sequential recovery provides its clearest benefit at 1.7B, where it
achieves the global best faithfulness score ($0.868$, $+0.007$ over
both-filtered for the same generator). This advantage does not
transfer to larger generators: for 4B and 8B, sequential recovery
is comparable to hard filtering on faithfulness and slightly lower
on ROUGE-L at 8B, suggesting that recovered samples from stronger
generators are structurally different from their native output.
Faithfulness decreases marginally with generator scale ($-0.007$
absolute, $1.7\text{B}{\to}8\text{B}$ under both-filtered), a
consistent but small trade-off against the lexical quality gains.
Full results across all five conditions are in
Table~\ref{tab:exp2c_full} (Appendix~\ref{app:adaptive}).

\section{Conclusion}
\label{sec:conclusion}

We presented a controlled study of LLM-as-judge curation pipelines
across gate configurations, recovery strategies, and generator scales.
Exact-provenance hallucination gating outperforms reward-only filtering
and post-hoc retrieved evidence, and hallucination and reward gates
reject largely disjoint failure populations, confirming both are
necessary. Reward-only scoring alone catastrophically over-rejects
faithful samples regardless of judge strength. Structured adaptive
repair outperforms naive retry on yield, recovery rate, and injection
recall across all generator sizes, with the injection recall advantage
growing to nearly double at 8B ($89.3\%$ vs $48.7\%$), and the
DiagnosticProbe's failure-mode telemetry surfaces diagnostic signal
that naive retry discards. Generator scale is the dominant quality
driver, though curation provides its clearest benefit when smaller
generators are non-negotiable, where exact-provenance gating with
adaptive recovery closes a meaningful fraction of the quality gap.
A genuine but small faithfulness trade-off with scale ($-0.007$
absolute) can be addressed using the HallucinationGate when source
fidelity is a hard requirement.
\section*{Limitations}

All generators, judges, and the fine-tuning base are from the Qwen3
family; cross-family replication is needed to confirm generality.
Provenance gating is validated on a single benchmark (FaithDial,
$n{=}3{,}507$). The downstream benefit of filtration conditions
diminishes with generator scale: within-generator condition spreads
($0.002$--$0.010$ ROUGE-L) are plausibly within noise at
$n{=}1{,}400$, suggesting that structured repair is most valuable
when baseline generator quality is low. Downstream fine-tuning uses
a single base model (Qwen3-4B with LoRA). The three-domain corpus
(legal, biomedical, Wikipedia) may not represent specialized or
low-resource domains. The DiagnosticProbe is inherently sequential,
as each sample requires up to six chained LLM calls whose outputs
determine the next step, and although the pipeline overlaps
independent samples via asyncio with bounded concurrency, per-sample
probe latency remains a bottleneck.

\section*{Ethics Statement}

This work uses LLM-as-judge systems as quality filters. Such judges
inherit biases from their training data; the quality of any curated
corpus depends on judge quality and should be validated against
ground-truth labels on domain-representative samples before deployment.
No new pretraining is performed; all inference uses existing publicly
available checkpoints. Data curation experiments ran on an NVIDIA A100
(80\,GB) and fine-tuning on an RTX Pro 6000 (96\,GB); total compute
was approximately 26 GPU-hours across both stages. All source corpora
are derived from publicly licensed datasets (CUAD: Apache~2.0,
PubMedQA: MIT, Wikipedia: CC~BY-SA, FaithDial: MIT) and contain no
personally identifiable information.


\bibliographystyle{unsrtnat}
\bibliography{references}

@misc{distilabel,
  title  = {distilabel: An AI Feedback (AIF) framework for building datasets with and for LLMs},
  author = {{Argilla}},
  year   = {2024},
  url    = {https://github.com/argilla-io/distilabel},
}

@inproceedings{penedo2024datatrove,
  title     = {{DataTrove}: Large Scale Data Processing},
  author    = {Penedo, Guilherme and others},
  booktitle = {NeurIPS Datasets and Benchmarks Track},
  year      = {2024},
}

@article{agentinstruct,
  title  = {{AgentInstruct}: Toward Generative Teaching with Agentic Flows},
  author = {Mitra, Arindam and others},
  year   = {2024},
  journal= {arXiv preprint arXiv:2407.03502},
}

@article{cui2023ultrafeedback,
  title  = {{UltraFeedback}: Boosting Language Models with High-Quality Feedback},
  author = {Cui, Ganqu and others},
  year   = {2023},
  journal= {arXiv preprint arXiv:2310.01377},
}

@inproceedings{maynez2020faithfulness,
  title     = {On Faithfulness and Factuality in Abstractive Summarization},
  author    = {Maynez, Joshua and others},
  booktitle = {ACL},
  year      = {2020},
}

@inproceedings{rashkin2021increasing,
  title     = {Increasing Faithfulness in Knowledge-Grounded Dialogue with Controllable Features},
  author    = {Rashkin, Hannah and others},
  booktitle = {ACL},
  year      = {2021},
}

@article{honovich2022true,
  title  = {{TRUE}: Re-evaluating Factual Consistency Evaluation},
  author = {Honovich, Or and others},
  year   = {2022},
  journal= {arXiv preprint arXiv:2204.04991},
}

@inproceedings{faithdial,
  title     = {{FaithDial}: A Faithful Benchmark for Information-Seeking Dialogue},
  author    = {Dziri, Nouha and others},
  booktitle = {Transactions of the Association for Computational Linguistics},
  year      = {2022},
}

@article{yuan2024selfrewarding,
  title  = {Self-Rewarding Language Models},
  author = {Yuan, Weizhe and others},
  year   = {2024},
  journal= {arXiv preprint arXiv:2401.10020},
}

@article{lee2023rlaif,
  title  = {{RLAIF}: Scaling Reinforcement Learning from Human Feedback with AI Feedback},
  author = {Lee, Harrison and others},
  year   = {2023},
  journal= {arXiv preprint arXiv:2309.00267},
}

@misc{nemo-data-designer,
  author = {The NeMo Data Designer Team, NVIDIA},
  title = {NeMo Data Designer: A framework for generating synthetic data from scratch or based on your own seed data},
  howpublished = {\url{https://github.com/NVIDIA-NeMo/DataDesigner}},
  year = {2025},
  note = {GitHub Repository},
}

@inproceedings{wang2023selfconsistency,
  title     = {Self-Consistency Improves Chain of Thought Reasoning in Language Models},
  author    = {Wang, Xuezhi and Wei, Jason and Schuurmans, Dale and Le, Quoc and Chi, Ed and Narang, Sharan and Chowdhery, Aakanksha and Zhou, Denny},
  booktitle = {ICLR},
  year      = {2023},
}

@inproceedings{zheng2024judging,
  title     = {Judging {LLM}-as-a-Judge with {MT}-Bench and Chatbot Arena},
  author    = {Zheng, Lianmin and Chiang, Wei-Lin and Sheng, Ying and Zhuang, Siyuan and Wu, Zhanghao and Zhuang, Yonghao and Lin, Zi and Li, Zhuohan and Li, Dacheng and Xing, Eric P. and Zhang, Hao and Gonzalez, Joseph E. and Stoica, Ion},
  booktitle = {NeurIPS Datasets and Benchmarks Track},
  year      = {2023},
}

@inproceedings{liu2023geval,
  title     = {{G-Eval}: {NLG} Evaluation using {GPT-4} with Better Human Alignment},
  author    = {Liu, Yang and Iter, Dan and Xu, Yichong and Wang, Shuohang and Xu, Ruochen and Zhu, Chenguang},
  booktitle = {EMNLP},
  year      = {2023},
}

@article{cohen2023lmvslm,
  title   = {{LM} vs {LM}: Detecting Factual Errors via Cross Examination},
  author  = {Cohen, Roi and Hamri, May and Geva, Mor and Globerson, Amir},
  journal = {arXiv preprint arXiv:2305.13281},
  year    = {2023},
}

@misc{qwen3_techreport,
  title        = {Qwen3 Technical Report},
  author       = {{Qwen Team}},
  year         = {2025},
  howpublished = {\url{https://huggingface.co/Qwen}},
}

@article{cuad,
  title   = {{CUAD}: An Expert-Annotated {NLP} Dataset for Legal Contract Review},
  author  = {Hendrycks, Dan and Burns, Collin and Chen, Anya and Ball, Spencer},
  journal = {NeurIPS Datasets and Benchmarks Track},
  year    = {2021},
}

@inproceedings{jin2019pubmedqa,
  title     = {{PubMedQA}: A Dataset for Biomedical Research Question Answering},
  author    = {Jin, Qiao and Dhingra, Bhuwan and Liu, Zhengping and Cohen, William W. and Lu, Xinghua},
  booktitle = {EMNLP},
  year      = {2019},
}

@inproceedings{hu2021lora,
  title     = {{LoRA}: Low-Rank Adaptation of Large Language Models},
  author    = {Hu, Edward J. and Shen, Yelong and Wallis, Phillip and Allen-Zhu, Zeyuan and Li, Yuanzhi and Wang, Shean and Wang, Lu and Chen, Weizhu},
  booktitle = {ICLR},
  year      = {2022},
}

@misc{unsloth,
  title        = {Unsloth: 2$\times$ faster, 50\% less memory {LLM} fine-tuning},
  author       = {{Unsloth AI}},
  year         = {2024},
  howpublished = {\url{https://github.com/unslothai/unsloth}},
}

@article{
dong2023raft,
title={{RAFT}: Reward rAnked FineTuning for Generative Foundation Model Alignment},
author={Hanze Dong and Wei Xiong and Deepanshu Goyal and Yihan Zhang and Winnie Chow and Rui Pan and Shizhe Diao and Jipeng Zhang and KaShun SHUM and Tong Zhang},
journal={Transactions on Machine Learning Research},
issn={2835-8856},
year={2023},
url={https://openreview.net/forum?id=m7p5O7zblY},
note={}
}

@article{
singh2024beyond,
title={Beyond Human Data: Scaling Self-Training for Problem-Solving with Language Models},
author={Avi Singh and John D Co-Reyes and Rishabh Agarwal and Ankesh Anand and Piyush Patil and Xavier Garcia and Peter J Liu and James Harrison and Jaehoon Lee and Kelvin Xu and Aaron T Parisi and Abhishek Kumar and Alexander A Alemi and Alex Rizkowsky and Azade Nova and Ben Adlam and Bernd Bohnet and Gamaleldin Fathy Elsayed and Hanie Sedghi and Igor Mordatch and Isabelle Simpson and Izzeddin Gur and Jasper Snoek and Jeffrey Pennington and Jiri Hron and Kathleen Kenealy and Kevin Swersky and Kshiteej Mahajan and Laura A Culp and Lechao Xiao and Maxwell Bileschi and Noah Constant and Roman Novak and Rosanne Liu and Tris Warkentin and Yamini Bansal and Ethan Dyer and Behnam Neyshabur and Jascha Sohl-Dickstein and Noah Fiedel},
journal={Transactions on Machine Learning Research},
issn={2835-8856},
year={2024},
url={https://openreview.net/forum?id=lNAyUngGFK},
note={Expert Certification}
}

@inproceedings{lee-etal-2025-checkeval,
    title = "{C}heck{E}val: A reliable {LLM}-as-a-Judge framework for evaluating text generation using checklists",
    author = "Lee, Yukyung  and
      Kim, JoongHoon  and
      Kim, Jaehee  and
      Cho, Hyowon  and
      Kang, Jaewook  and
      Kang, Pilsung  and
      Kim, Najoung",
    editor = "Christodoulopoulos, Christos  and
      Chakraborty, Tanmoy  and
      Rose, Carolyn  and
      Peng, Violet",
    booktitle = "Proceedings of the 2025 Conference on Empirical Methods in Natural Language Processing",
    month = nov,
    year = "2025",
    address = "Suzhou, China",
    publisher = "Association for Computational Linguistics",
    url = "https://aclanthology.org/2025.emnlp-main.796/",
    doi = "10.18653/v1/2025.emnlp-main.796",
    pages = "15771--15798",
    ISBN = "979-8-89176-332-6"
}

@inproceedings{vllm,
author = {Kwon, Woosuk and Li, Zhuohan and Zhuang, Siyuan and Sheng, Ying and Zheng, Lianmin and Yu, Cody Hao and Gonzalez, Joseph and Zhang, Hao and Stoica, Ion},
title = {Efficient Memory Management for Large Language Model Serving with PagedAttention},
year = {2023},
isbn = {9798400702297},
publisher = {Association for Computing Machinery},
address = {New York, NY, USA},
url = {https://doi.org/10.1145/3600006.3613165},
doi = {10.1145/3600006.3613165},
booktitle = {Proceedings of the 29th Symposium on Operating Systems Principles},
pages = {611–626},
numpages = {16},
location = {Koblenz, Germany},
series = {SOSP '23}
}

\appendix

\clearpage
\section{Related Work}
\label{app:related}

\textbf{LLM-as-judge.} \citet{zheng2024judging} establish LLM-as-judge as a workable proxy for
human preference at scale. \citet{liu2023geval} propose G-Eval, a
holistic chain-of-thought scoring approach. Structured claim
verification~\citep{cohen2023lmvslm} breaks a response into atomic
claims and verifies each against source evidence; we use this
formulation for our hallucination gate.

\textbf{Faithfulness in generated data.} Fluency and faithfulness are weakly correlated, motivating dedicated
checks~\citep{maynez2020faithfulness}. NLI-based post-hoc filters have
been proposed~\citep{rashkin2021increasing,honovich2022true}, and
FaithDial~\citep{faithdial} provides ground-truth faithfulness labels
used to validate gate calibration.

\textbf{Self-consistency and retry.} \citet{wang2023selfconsistency} show that sampling multiple completions
and aggregating improves chain-of-thought reasoning. Our naive retry
baseline is the minimal variant: regenerate and re-gate, no aggregation.

\textbf{AI feedback and self-improvement.} \citet{lee2023rlaif} show that AI feedback can substitute for human
preference labels at scale. \citet{yuan2024selfrewarding} extend this to
self-rewarding loops where the model scores its own outputs; our reward
gate plays an analogous role as an automated quality signal applied at
corpus construction time.

\textbf{Synthetic data frameworks.} distilabel \citep{distilabel}, DataTrove \citep{penedo2024datatrove},
NeMo Data Designer \citep{nemo-data-designer}, and AgentInstruct
\citep{agentinstruct} compose generation, scoring, and filtering;
rejection is typically terminal. Our study is the first to compare
structured repair and naive retry recovery strategies on top of such
pipelines in a controlled ablation.

\clearpage
\section{Extended Experimental Results}
\label{app:results}

\subsection{Gate Method Comparison: Full Results}
\label{app:exp1a}

Table~\ref{tab:exp1a_full} reports precision, recall, and $F_1$ for
all four gate configurations across all three judge sizes on FaithDial
($n{=}3{,}507$). The main paper reports only $F_1$ for the 14B and
35B judges; full metrics and the 27B judge are included here.

\begin{table}[htpb]
  \centering
  \footnotesize
  \caption{Full gate method comparison on FaithDial ($n{=}3{,}507$).
           Exact-provenance gating achieves the highest $F_1$ at 27B
           and 35B; retrieved provenance leads marginally at 14B.
           Reward-only scoring degrades with judge strength, confirming
           it cannot serve as a faithfulness filter at any judge size.}
  \label{tab:exp1a_full}
  \vspace{4pt}
  \setlength{\tabcolsep}{4pt}
  \begin{tabular}{lrrrrrrrrrr}
    \toprule
    & \multicolumn{3}{c}{\textbf{14B judge}}
    & \multicolumn{3}{c}{\textbf{27B judge}}
    & \multicolumn{3}{c}{\textbf{35B judge}} \\
    \cmidrule(lr){2-4}\cmidrule(lr){5-7}\cmidrule(lr){8-10}
    \textbf{Gate configuration}
      & $F_1$ & $P$ & $R$
      & $F_1$ & $P$ & $R$
      & $F_1$ & $P$ & $R$ \\
    \midrule
    Exact provenance
      & 0.494 & 0.357 & 0.803
      & 0.594 & 0.552 & 0.644
      & \textbf{0.614} & \textbf{0.588} & 0.642 \\
    Retrieved provenance
      & \textbf{0.514} & 0.388 & 0.759
      & 0.573 & 0.533 & 0.618
      & 0.590 & 0.566 & 0.615 \\
    Oracle holistic
      & 0.489 & 0.335 & \textbf{0.903}
      & 0.569 & 0.436 & 0.819
      & 0.599 & 0.479 & \textbf{0.799} \\
    Reward only
      & 0.291 & 0.214 & 0.454
      & 0.185 & 0.230 & 0.154
      & 0.184 & 0.272 & 0.140 \\
    \bottomrule
  \end{tabular}
\end{table}

Exact-provenance gating improves consistently with judge strength,
reaching $F_1{=}0.614$ at 35B. Retrieved provenance leads marginally
at 14B ($F_1{=}0.514$ vs $0.494$) but falls behind at 27B and 35B,
confirming that the benefit of exact provenance grows with judge
capacity. Oracle holistic scoring achieves the highest recall at 14B
($0.903$) but the lowest precision, making it unsuitable as a
high-precision curation gate despite its strong recall. Reward-only
scoring degrades with judge strength: $F_1$ falls from $0.291$ at
14B to $0.184$ at 35B as stronger judges apply a stricter reward
rubric that over-rejects faithful samples even more aggressively.

\subsection{Judge-Agnostic Provenance Transfer}
\label{app:exp1d}

Table~\ref{tab:exp1d} reports $F_1$, precision, recall, and inference
time for G-Eval and CheckEval scoring under exact and retrieved
provenance across all three judge sizes on FaithDial ($n{=}3{,}507$). This experiment tests whether the faithfulness
detection signal from exact-provenance gating transfers across judge
families, or whether it is an artefact of a specific model.

\begin{table}[htpb]
  \centering
  \footnotesize
  \caption{Judge-agnostic provenance transfer on FaithDial
           ($n{=}3{,}507$, threshold $= 0.5$). CheckEval with exact
           provenance achieves the highest $F_1$ at 27B and 35B;
           G-Eval achieves high recall but low precision across all
           judges. Bold = best $F_1$ per judge.}
  \label{tab:exp1d}
  \vspace{4pt}
  \setlength{\tabcolsep}{3pt}
  \begin{tabular}{llcccc}
    \toprule
    \textbf{Judge} & \textbf{Condition} & $F_1$ & $P$ & $R$ & \textbf{Time} \\
    \midrule
    14B & geval, retrieved     & 0.443 & 0.298 & 0.865 & 140.8 \\
    14B & geval, exact         & 0.449 & 0.291 & 0.981 & 109.4 \\
    14B & checkeval, retrieved & 0.513 & 0.372 & 0.831 & 359.7 \\
    14B & \textbf{checkeval, exact} & \textbf{0.568} & 0.424 & 0.857 & 162.3 \\
    \midrule
    27B & geval, retrieved     & 0.520 & 0.362 & 0.920 & 161.5 \\
    27B & geval, exact         & 0.535 & 0.371 & 0.960 & 154.8 \\
    27B & checkeval, retrieved & 0.662 & 0.690 & 0.635 & 256.4 \\
    27B & \textbf{checkeval, exact} & \textbf{0.686} & 0.689 & 0.683 & 254.8 \\
    \midrule
    35B & geval, retrieved     & 0.485 & 0.331 & 0.906 & 153.8 \\
    35B & geval, exact         & 0.489 & 0.331 & 0.937 & 139.3 \\
    35B & checkeval, retrieved & 0.651 & 0.699 & 0.610 & 229.3 \\
    35B & \textbf{checkeval, exact} & \textbf{0.672} & 0.695 & 0.651 & 214.1 \\
    \bottomrule
  \end{tabular}
\end{table}

CheckEval with exact provenance is the best condition at every judge
size, outperforming retrieved provenance by $+0.021$-$+0.055$ $F_1$.
G-Eval achieves high recall ($0.865$-$0.981$) but poor precision,
making it unsuitable as a curation gate. Beyond detection quality,
exact provenance eliminates retrieval overhead entirely: under the
14B judge, switching from retrieved to exact reduces inference time
by $197$s while improving recall by $+0.026$. The consistent $F_1$
advantage of exact over retrieved provenance across all three judges
confirms the finding is not judge-specific.

\subsection{Gate Overlap: Full Results}
\label{app:gate_overlap_14b}

Table~\ref{tab:gate_overlap_full} reports gate rejection-set overlap
for both judges across all three generators. The main paper reports
35B judge numbers only; the 14B judge is included here to confirm
that gate complementarity holds across judge strengths.

\begin{table}[htpb]
  \centering
  \footnotesize
  \caption{Gate rejection-set overlap, 14B and 35B judges. Jaccard
           remains below $0.33$ across all conditions; 35B applies a
           substantially stricter reward gate.}
  \label{tab:gate_overlap_full}
  \vspace{4pt}
  \setlength{\tabcolsep}{3pt}
  \begin{tabular}{llccccc}
    \toprule
    & & & \multicolumn{3}{c}{\textit{Rejected (\% of corpus)}}
        & \textit{Accepted} \\
    \cmidrule(lr){4-6}\cmidrule(lr){7-7}
    \textbf{Judge} & \textbf{Gen.} & \textbf{Jacc.}
    & \textbf{Hall} & \textbf{Rew.} & \textbf{Both}
    & \textbf{Yield (\%)} \\
    \midrule
    14B & 1.7B & 0.23  & 20.5 & 22.2 &  9.6 & 57.9 \\
    14B & 4B   & 0.31  & 11.9 & 17.2 & 13.4 & 57.5 \\
    14B & 8B   & 0.28  & 13.4 & 16.9 & 11.8 & 57.9 \\
    \midrule
    35B & 1.7B & 0.317 &  7.3 & 32.3 & 18.4 & 42.1 \\
    35B & 4B   & 0.324 &  6.5 & 38.6 & 21.6 & 33.3 \\
    35B & 8B   & 0.284 &  8.5 & 36.6 & 17.8 & 37.1 \\
    \bottomrule
  \end{tabular}
\end{table}

Gate complementarity is consistent across both judges: Jaccard
remains below $0.33$ in all six conditions. The 35B judge applies a
substantially stricter reward gate, with reward-only rejection rising
from $17$-$22\%$ under 14B to $33$-$39\%$ under 35B, while the
hallucination gate becomes only marginally stricter ($-2$-$7$
percentage points). This asymmetry confirms that judge strength
primarily affects reward sensitivity rather than hallucination
detection, and that the two gates remain structurally complementary
regardless of the operating judge.

\subsection{Injection Recall by Failure Type}
\label{app:injection_recall}

\begin{table}[htpb]
  \centering
  \footnotesize
  \caption{Injection recall by failure type, 14B and 35B judges,
           gate-only (no recovery). \emph{instruction\_quality} is
           the hardest type across all conditions; \emph{contradicts\_source}
           and \emph{domain\_mismatch} are caught most reliably for
           4B and 8B generators.}
  \label{tab:gate_recall}
  \vspace{4pt}
  \setlength{\tabcolsep}{4pt}
  \begin{tabular}{llrrrr}
    \toprule
    \textbf{Judge} & \textbf{Gen.}
    & \textbf{contr.\ src}
    & \textbf{param.\ drift}
    & \textbf{dom.\ mismatch}
    & \textbf{instruct.\ qual.} \\
    \midrule
    14B & 1.7B & 20.3 & 17.3 & 26.7 & 23.8 \\
    14B & 4B   & 82.3 & 47.6 & 85.8 & 34.2 \\
    14B & 8B   & 89.0 & 77.1 & 88.0 & 42.4 \\
    \midrule
    35B & 1.7B & 29.5 & 34.0 & 37.3 & 26.3 \\
    35B & 4B   & 83.1 & 49.5 & 82.1 & 31.3 \\
    35B & 8B   & 89.9 & 78.4 & 85.4 & 28.1 \\
    \bottomrule
  \end{tabular}
\end{table}

\emph{instruction\_quality} is the hardest failure type at every
judge size and generator ($23$-$42\%$ recall), as it requires
reasoning about instruction intent rather than direct source
grounding. \emph{contradicts\_source} and \emph{domain\_mismatch}
are caught most reliably for 4B and 8B ($82$-$90\%$).
\emph{parametric\_drift} recall improves markedly with generator
size, suggesting larger generators produce more detectable
parametric confabulations.

\subsection{Adaptive Recovery: Stage Breakdown}
\label{app:adaptive}

Table~\ref{tab:exp2a} traces the adaptive pipeline stage by stage
for all three generators under the 14B judge.

\begin{table}[t]
  \centering
  \footnotesize
  \caption{Adaptive pipeline stage breakdown (14B judge). Total
    accepted: $19{,}940$ vs.\ $14{,}047$ for hard filtering
    ($+42\%$).}
  \label{tab:exp2a}
  \vspace{4pt}
  \setlength{\tabcolsep}{4pt}
  \begin{tabular}{lrrr}
    \toprule
    \textbf{Stage} & \textbf{1.7B} & \textbf{4B} & \textbf{8B} \\
    \midrule
    Raw input              & 7{,}812 & 8{,}433 & 8{,}039 \\
    HallGate pass          & 6{,}266 & 6{,}348 & 6{,}046 \\
    \quad Probe recovers   & 1{,}171 & 1{,}906 & 1{,}839 \\
    Probe$\!\to\!$Reward pass & 568  & 1{,}044 &   910   \\
    Reward main reject     & 1{,}741 & 1{,}475 & 1{,}397 \\
    \quad Refiner recovers &   501   & 1{,}315 & 1{,}555 \\
    \midrule
    \textbf{Total accepted} & \textbf{5{,}594} & \textbf{7{,}232} & \textbf{7{,}114} \\
    \textbf{Pass rate}      & 71.6\%  & 85.8\%  & 88.5\%  \\
    Hard-filter baseline    & 4{,}525 & 4{,}873 & 4{,}649 \\
    Yield gain              & +23.6\% & +48.4\% & +53.1\% \\
    \bottomrule
  \end{tabular}
\end{table}

Approximately $45$-$51\%$ of probe-recovered samples subsequently
fail the reward gate, explaining the gap between probe recovery rate
and final accepted count. Under the 35B judge (1.7B generator only),
the pipeline accepts $4{,}635$ samples ($59.3\%$ pass rate) with a
$23.6\%$ recovery rate and $69.0\%$ injection recall, confirming the
adaptive advantage holds under a stronger judge at lower overall
yield due to the stricter reward gate.

\begin{figure}[htbp]
  \centering
  \includegraphics[width=0.65\textwidth]{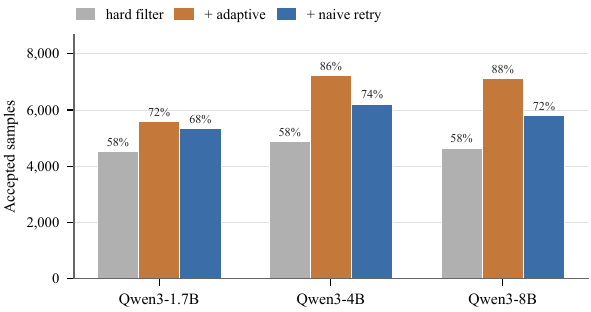}
  \caption{Accepted samples per generator under hard filtering,
    adaptive recovery, and naive retry.}
  \label{fig:yield}
\end{figure}

Table~\ref{tab:exp2c_full} reports downstream fine-tuning results
across all five curation conditions for completeness.

\begin{table}[htpb]
  \centering
  \footnotesize
  \caption{Full downstream fine-tuning results (Qwen3-4B base, LoRA,
    $n{=}1{,}400$ test set). Bold = best per generator per metric.}
  \label{tab:exp2c_full}
  \vspace{4pt}
  \setlength{\tabcolsep}{4pt}
  \begin{tabular}{llccc}
    \toprule
    \textbf{Gen.} & \textbf{Condition} & \textbf{R-L} & \textbf{BS-F1} & \textbf{Faith.} \\
    \midrule
    1.7B & unfiltered  & 0.5227 & 0.9059 & 0.8537 \\
    1.7B & hall        & 0.5372 & 0.9112 & 0.8581 \\
    1.7B & reward      & 0.5368 & 0.9116 & 0.8595 \\
    1.7B & both        & \textbf{0.5388} & 0.9112 & 0.8606 \\
    1.7B & sequential  & 0.5382 & \textbf{0.9113} & \textbf{0.8680} \\
    \midrule
    4B   & unfiltered  & 0.5577 & 0.9185 & 0.8542 \\
    4B   & hall        & 0.5535 & 0.9185 & 0.8546 \\
    4B   & reward      & 0.5599 & 0.9191 & 0.8548 \\
    4B   & both        & \textbf{0.5606} & 0.9197 & \textbf{0.8570} \\
    4B   & sequential  & 0.5604 & \textbf{0.9223} & 0.8563 \\
    \midrule
    8B   & unfiltered  & 0.5734 & 0.9187 & 0.8527 \\
    8B   & hall        & 0.5761 & 0.9184 & 0.8526 \\
    8B   & reward      & 0.5667 & 0.9175 & 0.8533 \\
    8B   & both        & \textbf{0.5764} & \textbf{0.9191} & \textbf{0.8540} \\
    8B   & sequential  & 0.5585 & 0.9189 & 0.8537 \\
    \bottomrule
  \end{tabular}
\end{table}

\clearpage
\section{DiagnosticProbe: Failure-Mode Routing}
\label{app:probe}

When a sample fails the HallucinationGate, the DiagnosticProbe
attempts recovery through two probe types, routed by the grounding
score returned by the gate.

\paragraph{Routing.} If the grounding score is below $0.5$, the
failure is likely a prompt-level grounding issue; Probe~2a runs
first, followed by Probe~1, then Probe~2b. If the grounding score
is at or above $0.5$, the failure is likely stochastic; Probe~1
runs first, followed by Probe~2. The first attempt that clears the
HallucinationGate is accepted and forwarded to the RewardGate. If
all attempts fail, the sample is permanently rejected.

\paragraph{Probe 1: Temperature sweep.} The sample is regenerated
at three temperatures in sequence: $0.3$, $0.7$, and $1.1$. This
targets stochastic failures where the original generation drifted
at high temperature; lower temperatures encourage the model to
stay closer to the source chunk. The first passing attempt wins.

\paragraph{Probe 2: Prompt variants.} Two sub-variants are applied
in sequence. Probe~2a uses a strict source-anchored prompt that
explicitly constrains the model to cite only from the provided
source chunk, targeting parametric leakage. Probe~2b regenerates
the instruction from scratch using the default prompt, targeting
cases where the original instruction formulation was the source of
the failure rather than the answer generation.

\paragraph{RewardRefiner.} When a sample fails the RewardGate, the
RewardRefiner issues a single targeted rewrite call. The judge
returns the worst-scoring evaluation axis alongside a structured
critique. The refiner prompt is:

\begin{quote}
\small
\textit{The following answer was scored low on \{axis\}. Judge
feedback: ``\{weakness\}''. Rewrite the answer to specifically
improve \{axis\}.}
\end{quote}

The rewritten sample re-enters the RewardGate once. If it fails
again it is permanently rejected. The refiner does not alter the
underlying factual claims, only the quality dimensions flagged by
the judge.

\clearpage
\section{Discussion}
\label{app:discussion}

\textbf{Downstream analysis.}
Within any generator block, ROUGE-L spread is at most
$0.002$/$0.007$/$0.010$ (1.7B/4B/8B) and faithfulness spread
$\leq\!0.003$, small relative to the cross-generator effect.
\emph{Both}-filtered leads or ties on ROUGE-L and faithfulness in
every slot, though margins are never practically meaningful.
There is a genuine faithfulness-ROUGE trade-off with scale:
moving from 1.7B to 8B gains $+0.036$ ROUGE-L while losing
$-0.007$ faithfulness, consistently across all conditions. The
8:1 ratio strongly favours the quality gain in most settings;
practitioners for whom source fidelity is a hard requirement can
apply the HallucinationGate explicitly at any generator size.

\textbf{Practical recommendations.}
Gate with both hallucination and reward signals using exact source
provenance and recover failures with the DiagnosticProbe and
RewardRefiner. Generator scale is the dominant quality driver, but
curation is most valuable precisely where larger generators are
non-negotiable due to latency, memory, or edge deployment
constraints. In those settings the curation pipeline closes a
meaningful fraction of the quality gap that generator capacity
cannot. The DiagnosticProbe's failure-mode telemetry, covering
temperature drift, parametric leakage, and instruction quality
failures, is independently useful for prompt engineering regardless
of downstream quality gains.

\textbf{Why adaptive repair outperforms naive retry.}
Targeted diagnosis applies mode-specific patches rather than blind
resampling. Structured repair correctly fails to fix genuine
injections, since patching a hallucination requires altering
content rather than style, driving injection recall $+13.5$ points
above naive retry. The staged probe-then-reward pathway provides
two targeted recovery attempts per hall-rejected sample, followed
by orthogonal reward refining. Together these yield a $32.5\%$
recovery rate vs $24.9\%$ for naive retry at 1.7B, with the
advantage growing to $+38.1$ points at 8B. That $45$-$51\%$ of
probe-recovered samples subsequently fail the reward gate confirms
hallucination and reward failures are largely orthogonal,
consistent with Section~\ref{sec:gates}.

\textbf{When naive retry is a reasonable choice.}
Naive retry is viable when simplicity and throughput matter more
than maximising injection recall. It requires no failure-mode
taxonomy and only two extra gate calls per attempt. The yield gap
($68.4\%$ vs $71.6\%$ at 1.7B) is modest, and its lower natural
rejection rate is a genuine benefit when preserving non-injected
samples is the primary concern. The tradeoff shifts further against
naive retry at larger generator sizes, where adaptive simultaneously
achieves lower natural rejection and higher injection recall.

\textbf{Why filtration is flat downstream.}
At approximately 3,000 training samples per condition, quality
differences between filtration strategies are likely washed out by
sample size, and all conditions already remove the noisiest samples.
This flatness may not hold at smaller training sizes or in
harder-domain settings where source faithfulness is more
consequential.

\textbf{Scope of generalisation.}
All generators, judges, and the fine-tuning base are from the Qwen3
family. Whether the DiagnosticProbe's failure-mode taxonomy
transfers across generator families is an open question, and the
downstream flatness may partly reflect Qwen3's pre-training coverage
reducing sensitivity to curation at the scales we test.

\clearpage
\section{Hyperparameters}
\label{app:hyperparams}

\begin{table}[h]
  \centering
  \footnotesize
  \caption{Fine-tuning, pipeline, evaluation hyperparameters, compute budget,
    and software versions.}
  \label{tab:hparams}
  \vspace{4pt}
  \setlength{\tabcolsep}{4pt}
  \begin{tabular}{ll}
    \toprule
    \textbf{Component} & \textbf{Value} \\
    \midrule
    \multicolumn{2}{l}{\textit{Fine-tuning}} \\
    Base model          & Qwen3-4B-Instruct-2507 \\
    LoRA rank / $\alpha$ & 16 / 32 \\
    Optimizer / lr      & AdamW / $2\!\times\!10^{-4}$ \\
    Effective batch     & 32 \\
    Epochs / patience   & 3 / 2 \\
    Eval interval       & 0.5 epoch \\
    Train / val (hall)  & 5{,}000 / 500 \\
    Train / val (other) & ${\sim}3{,}000$ / 500 \\
    \midrule
    \multicolumn{2}{l}{\textit{Pipeline}} \\
    Gate thresholds     & hall $0.8$, reward $0.7$ \\
    Retry budget        & 5 attempts \\
    Decoding temp.\ (gen) & $0.7$ \\
    Embedding retriever & \texttt{BAAI/bge-base-en-v1.5} \\
    \midrule
    \multicolumn{2}{l}{\textit{Compute: Data Curation}} \\
    GPU                 & A100 (80\,GB) $\times 1$ \\
    Total GPU-hours     & ${\sim}$18 hours \\
    \midrule
    \multicolumn{2}{l}{\textit{Compute: fine-tuning and eval}} \\
    GPU                 & RTX Pro 6000 (96\,GB)  $\times 1$ \\
    Total GPU-hours     & 8 hours \\
    \midrule
    \multicolumn{2}{l}{\textit{Software versions}} \\
    Unsloth & v0.1.41-beta \citep{unsloth} \\
    vLLM & vllm 0.21.0 \citep{vllm} \\
    LiteLLM             & $\geq$1.40 \\
    \texttt{openai}     & $\geq$1.30 \\
    \texttt{sentence-transformers} & $\geq$2.2 \\
    \texttt{datasets}   & $\geq$2.14 \\
    \texttt{rouge\_score} & 0.1.2 \\
    \texttt{bert\_score} & 0.3.13 (\texttt{roberta-large}) \\
    \bottomrule
  \end{tabular}
\end{table}

\end{document}